\documentclass[9pt,conference]{IEEEtran}
\IEEEoverridecommandlockouts
\usepackage{cite}
\usepackage{amsmath,amssymb,amsfonts}
\usepackage{algorithmic}
\usepackage{graphicx}
\usepackage{textcomp}
\usepackage{booktabs}
\usepackage{multirow}
\usepackage{subfigure}
\usepackage{url}
\usepackage{xcolor}
\def\BibTeX{{\rm B\kern-.05em{\sc i\kern-.025em b}\kern-.08em
    T\kern-.1667em\lower.7ex\hbox{E}\kern-.125emX}}
\begin{document}

% \title{Conference Paper Title*\\
% {\footnotesize \textsuperscript{*}Note: Sub-titles are not captured for https://ieeexplore.ieee.org  and
\title{Norm Augmented Graph AutoEncoders for Link Prediction}

\author{Yunhui~Liu$^{\dag\star}$,
        Huaisong~Zhang$^{\ddag\star}$,
        Xinyi~Gao$^{\S}$,
        Liuye~Guo$^{\dag}$,
        Zhen~Tao$^{\dag}$,
        Tieke~He$^{\dag*}$\\
        $^{\dag}$State Key Laboratory for Novel Software Technology, Nanjing University, Nanjing, China\\
        $^{\ddag}$Tsinghua Shenzhen International Graduate School, Tsinghua University, Shenzhen, China\\
        $^{\S}$School of Electrical Engineering and Computer Science, The University of Queensland, Brisbane, Australia
        \thanks{$^{\star}$Equal contribution. $^*$Corresponding author (hetieke@gmail.com).}}

% \author{1\textsuperscript{st}\IEEEauthorblockN{Yunhui Liu$^{\star}$}
% \IEEEauthorblockA{
% \textit{State Key Laboratory for} \\
% \textit{Novel Software Technology} \\
% \textit{Nanjing University}\\
% Nanjing, China \\
% lyhcloudy1225@gmail.com}

% \and
% \IEEEauthorblockN{2\textsuperscript{nd} Huaisong Zhang$^{\star}$}
% \IEEEauthorblockA{
% % \textit{Tsinghua Shenzhen International Graduate School} \\
% \textit{Tsinghua University}\\
% Shenzhen, China \\
% nianbai006@gmail.com}

% \and
% \IEEEauthorblockN{3\textsuperscript{rd} Xinyi Gao}
% \IEEEauthorblockA{
% % \textit{School of Electrical Engineering and Computer Science} \\
% \textit{The University of Queensland}\\
% Brisbane, Australia \\
% xinyi.gao@uq.edu.au}
% \and
% \IEEEauthorblockN{4\textsuperscript{th} Liuye Guo}
% \IEEEauthorblockA{
% \textit{State Key Laboratory for} \\
% \textit{Novel Software Technology} \\
% \textit{Nanjing University}\\
% Nanjing, China \\
% liuye.guo1@gmail.com}
% \and
% \IEEEauthorblockN{5\textsuperscript{th} Zhen Tao}
% \IEEEauthorblockA{
% \textit{State Key Laboratory for} \\
% \textit{Novel Software Technology} \\
% \textit{Nanjing University}\\
% Nanjing, China \\
% zhentao@smail.nju.edu.cn}
% \and
% \IEEEauthorblockN{6\textsuperscript{th} Tieke He$^*$}
% \IEEEauthorblockA{
% \textit{State Key Laboratory for} \\
% \textit{Novel Software Technology} \\
% \textit{Nanjing University}\\
% Nanjing, China \\
% hetieke@gmail.com}
% \thanks{$^*$Corresponding author.}
% }

\maketitle

\begin{abstract}
Link Prediction (LP) is a crucial problem in graph-structured data. Graph Neural Networks (GNNs) have gained prominence in LP, with Graph AutoEncoders (GAEs) being a notable representation. 
However, our empirical findings reveal that GAEs' LP performance suffers heavily from the long-tailed node degree distribution, i.e., low-degree nodes tend to exhibit inferior LP performance compared to high-degree nodes.
\emph{What causes this degree-related bias, and how can it be mitigated?} 
In this study, we demonstrate that the norm of node embeddings learned by GAEs exhibits variation among nodes with different degrees, underscoring its central significance in influencing the final performance of LP.
Specifically, embeddings with larger norms tend to guide the decoder towards predicting higher scores for positive links and lower scores for negative links, thereby contributing to superior performance.
This observation motivates us to improve GAEs' LP performance on low-degree nodes by increasing their embedding norms, which can be implemented simply yet effectively by introducing additional self-loops into the training objective for low-degree nodes. This norm augmentation strategy can be seamlessly integrated into existing GAE methods with light computational cost. Extensive experiments on various datasets and GAE methods show the superior performance of norm-augmented GAEs.
\end{abstract}


\section{Introduction}
Link prediction (LP) which aims to predict missing or newly forming links within an observed graph, has been adopted in many applications across different domains such as social networks \cite{xGCN}, recommender systems \cite{SHaRe}, knowledge graphs \cite{DEKG-ILP}, and drug discovery \cite{abbas2021application}. 
% Traditional methods for this link prediction problem involve heuristics based on neighborhood statistics, such as predicting the existence of links whose endpoints have many neighbors in common, or distances involving paths between nodes \cite{LPSurvey}.
Recently, with the rising popularity of graph neural networks (GNNs), several GNN-based methods \cite{GAE, SEAL, Neo-GNNs, PS2} have achieved state-of-the-art performance on LP tasks.

% Graph neural networks (GNNs) have become widely adopted for machine learning tasks that deal with graph-structured data, consistently demonstrating remarkable performance across a variety of applications and downstream tasks \cite{GCN, SAGE, GAT, MCond, SEAL}. Among these tasks, link prediction (LP) holds particular significance as it aims to predict missing or newly forming links within an observed graph. Its applications are broad, spanning domains such as social networks \cite{xGCN}, recommender systems \cite{SHaRe}, knowledge graphs \cite{DEKG-ILP}, and drug discovery \cite{abbas2021application}. Traditional methods for LP have involved heuristics based on neighborhood statistics—such as the likelihood of link formation between endpoints with many neighbors in common—or involving path distances between nodes \cite{LPSurvey}.
% Recently, with the rising popularity of GNNs, several GNN-based methods \cite{GAE, SEAL, Neo-GNNs, PS2} have achieved state-of-the-art performance on LP tasks.

A prominent branch of GNNs for LP comprises Graph AutoEncoders (GAEs) \cite{GAE}. Typically, GAEs utilize message-passing neural networks such as GCN \cite{GCN} as the encoder to generate node embeddings, followed by a decoder such as the inner product to predict the probability of forming links between node pairs. These models employ binary cross-entropy loss, treating node pairs with observed links as positive samples, and those without as negative samples. While existing works exploit advanced message-passing encoders \cite{LGAE, GNAE, DGAE} or elaborate prior distribution assumptions \cite{ARGA, S-VAE, MSVGAE}, the inherent characteristics within graph data, such as node degree, are often not given adequate consideration, which may affect the performance.

In this study, we investigate the learned embeddings of GAEs and identify a degree bias issue; that is, GAEs demonstrate superior performance on high-degree nodes compared to low-degree nodes (see Section \ref{Sec: Performance Varying Node Degree}). Such a degree-related bias presents a significant challenge, especially given the prevalence of long-tailed distributions in the node degrees of real-world graphs—namely, only a few nodes have high degrees while most possess low degrees, or are even isolated with zero links \cite{albert2002statistical}. In view of this, a critical question emerges for GAEs in LP: \emph{what underlying factors contribute to this degree-related bias in LP, and how can we effectively counteract it?} Considerable works \cite{SL-DSGCN, Tail-GNN, ColdBrew, DegFairGNN} have focused on degree bias in node classification. They underscore that GNNs encounter difficulties with low-degree nodes due to the scarce neighborhood information, and that predictions for high-degree nodes are more likely to be influenced by training nodes because of their more extensive linkages with other nodes.
% Substantial researches \cite{SL-DSGCN, Tail-GNN, DegFairGNN} have concentrated on the degree bias in node classification, and they highlight that most GNNs adopt a message-passing approach, wherein information is iteratively aggregated from neighborhood to update node embeddings. Therefore, GNNs struggle with low-degree nodes due to the limited amount of available neighborhood information. Moreover, test predictions for high-degree nodes are more likely to be influenced by training nodes because they have more links with other nodes. 
Although degree bias in node classification has been extensively studied, LP remains relatively under-explored and presents unique challenges. We contend that the inherent imbalance in the training objectives of LP significantly amplifies degree-related performance disparities (see Section \ref{Sec: Embedding Norm Varying Node Degree}). Specifically, this imbalance causes high-degree nodes to have embeddings with larger norms. Consequently, larger-norm embeddings tend to lead the decoder to assign higher probabilities to positive samples and lower probabilities to negative samples, resulting in higher performance scores.

Therefore, we attempt to alleviate the degree-related bias in GAEs by increasing the embedding norms of nodes with lower degrees. We propose a simple yet effective strategy: introducing additional self-loops as positive samples into the training objective for nodes with lower degrees. This method leverages the principle that the inner product of the same embedding equates to the square of its norm. The number of supplementary self-loops assigned to a low-degree node is equal to the difference between the predefined degree threshold and its actual degree. Furthermore, our embedding \textbf{N}orm \textbf{A}ugmentation (\textbf{NA}) strategy can be combined with existing GAE models in a plug-and-play manner with minimal additional computational costs. Extensive experiments reveal that our norm augmentation strategy yields significant performance improvements when applied to various GAE models.

\section{Preliminary}
\subsection{Problem Definition}
Consider an attributed graph $\mathcal{G} = (\mathcal{V}, \mathcal{E}, \boldsymbol{X})$, where $\mathcal{V}=\{ v_i \}_{i=1}^{n}$ is the set of $n$ nodes and $\mathcal{E} \subset \mathcal{V} \times \mathcal{V}$ is the set of links with $e_{ij}$ denoting the link between the node $v_i$ and $v_j$, and $\boldsymbol{X} \in \mathbb{R}^{n \times d}$ represents the node attribute matrix. $\mathcal{N}_i = \{v_j|e_{ij} \in \mathcal{E}\}$ denotes the set of neighbors of node $v_i$, and the degree of $v_i$ is $d_i = |\mathcal{N}_i|$. 
The adjacency matrix of the graph is denoted as $\boldsymbol{A} \in \{0,1\}^{n \times n}$, where $\boldsymbol{A}_{ij} = 1$ if a link exists between node $v_i$ and $v_j$, and $\boldsymbol{A}_{ij} = 0$ otherwise. 
The link prediction problem is defined as follows. Given a complete graph
$\mathcal{G} = (\mathcal{V}, \mathcal{E}^c, \boldsymbol{X})$, where $\mathcal{E}^c$ is the set of all true links. However, the complete graph in real-world scenarios is typically partially observed, resulting in an incomplete link set denoted as $\mathcal{E}^+$. The goal of LP is to train a link predictor that takes $\mathcal{V}$, $\mathcal{E}^+$, and $\boldsymbol{X}$ as input to estimate the probability of whether an unobserved link belongs to the complete link set $\mathcal{E}^c$.

\subsection{GAEs for Link Prediction}
GAEs adopt the classic encoder-decoder framework, where a GNN encoder learns node embeddings and the decoder predicts the link existence probabilities given each pair of node embeddings. They are trained with the following binary cross-entropy loss:
\begin{equation}
\begin{aligned}
 & \mathcal{L}^+ = \frac{1}{|\mathcal{E}^+|}\sum_{(v_i, v_j)\in \mathcal{E}^+}\log \sigma(\boldsymbol{z}_i^\top \boldsymbol{z}_j),                                     \\
 & \mathcal{L}^- =  \frac{1}{|\mathcal{E}^-|}\sum_{(v_{i^\prime}, v_{j^\prime})\in \mathcal{E}^-}\log(1-\sigma(\boldsymbol{z}_{i^\prime}^\top \boldsymbol{z}_{j^\prime})), \\
 & \mathcal{L}_\text{GAEs} = - \left(\mathcal{L}^++\mathcal{L}^-\right),
\end{aligned}
\end{equation}
where $\boldsymbol{z}$ is the node embedding obtained from an encoder GNN and $\sigma$ is the Sigmoid function. We adopt the most widely used inner product decoder \cite{GAE, ARGA, LGAE, GNAE, DGAE, MSVGAE}. 
$\mathcal{E}^+$ is a set of positive samples, i.e., all observed links, and $\mathcal{E}^-$ is a set of negative samples, i.e., unobserved links. Typically, $\mathcal{E}^-$ comprises $|\mathcal{E}^+|$ negative samples randomly sampled from the entire unobserved link set $(\mathcal{V} \times \mathcal{V}) \setminus \mathcal{E}^+$ to speed up model training.

\section{Empirical Analysis}
In this section, we conduct empirical analysis on real-world graphs to show their properties for LP and the issue of GAEs on these datasets. This preliminary analysis lays a solid foundation and paves us a way to design better GAEs. We choose three widely used datasets Cora, CiteSeer, and CoraFull to perform the analysis. For data splits, we follow the experimental protocol as in \cite{LLP}, which splits the positive edges into three parts, 15\% for testing, 5\% for validation, and the rest for training.

\subsection{Degree Distribution}
\begin{figure}[h]
    \centering
    \subfigure[Cora]{\includegraphics[width=0.325\linewidth]{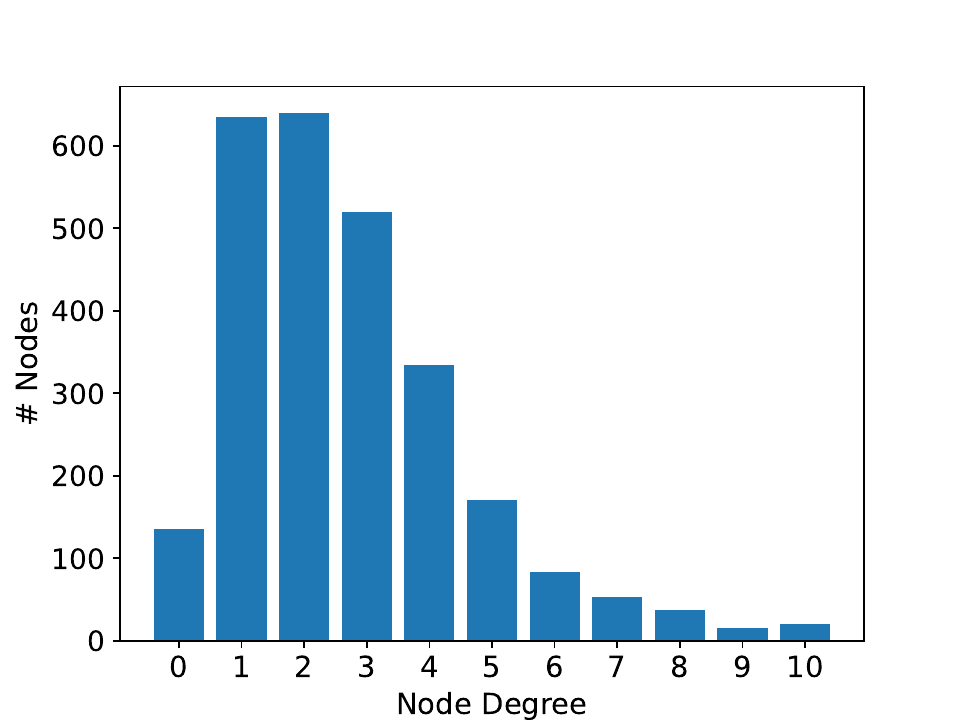}}  
    \subfigure[CiteSeer]{\includegraphics[width=0.325\linewidth]{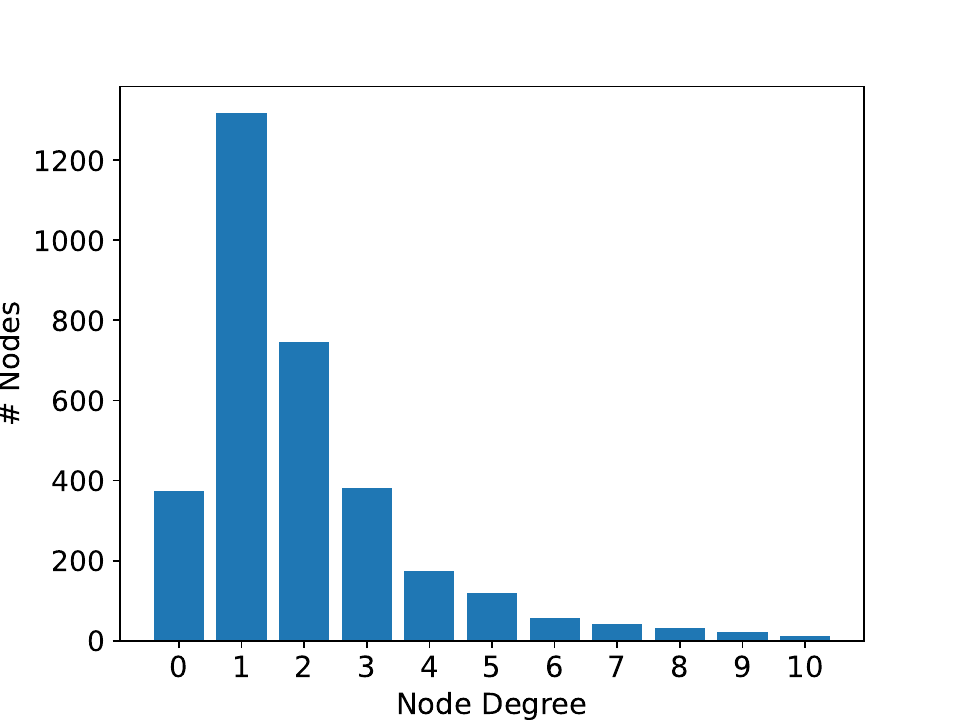}}
    \subfigure[CoraFull]{\includegraphics[width=0.325\linewidth]{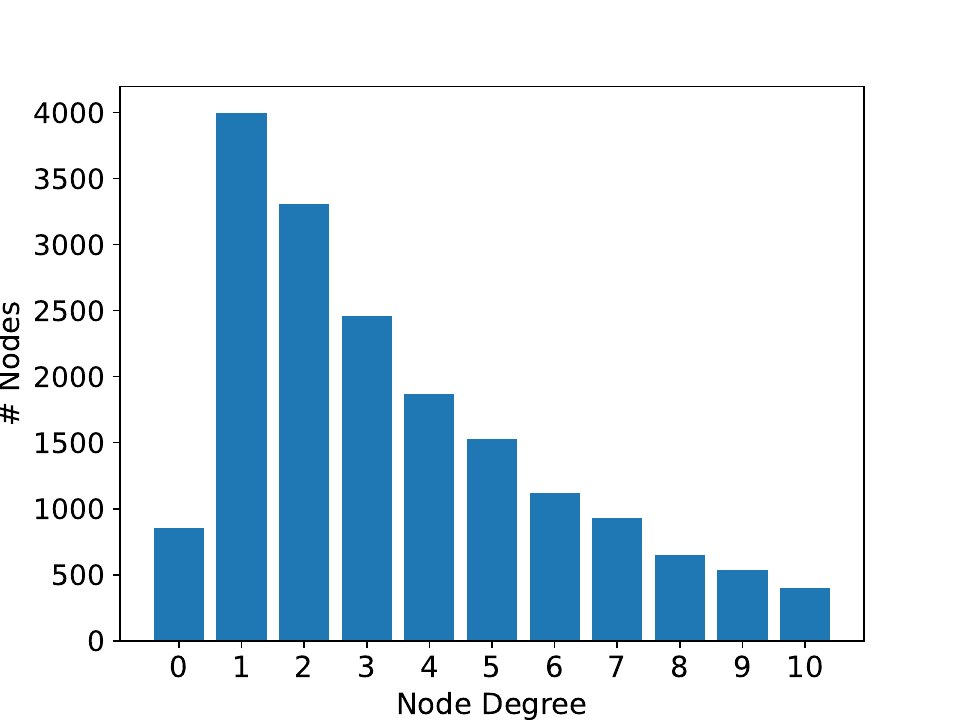}}
    \caption{Degree distribution of three datasets.}
    \label{Fig: Degree Distribution}
\end{figure}

Real-world graphs often follow a long-tailed distribution regarding node degrees \cite{albert2002statistical}. To verify this, we plot the degree distribution of the three datasets in Fig. \ref{Fig: Degree Distribution}. As we can see, only a few nodes are with high degrees, while the majority are low-degree nodes, and some are even isolated nodes with zero neighbors. Prior studies have noted a discrepancy in node classification accuracy among different node degrees, where the overall performance of GNNs is biased towards nodes with high degrees \cite{SL-DSGCN, Tail-GNN, ResNorm}.

\subsection{Performance Varying Node Degree}\label{Sec: Performance Varying Node Degree}
\begin{figure}[h]
    \centering
    \subfigure[Cora]{\includegraphics[width=0.325\linewidth]{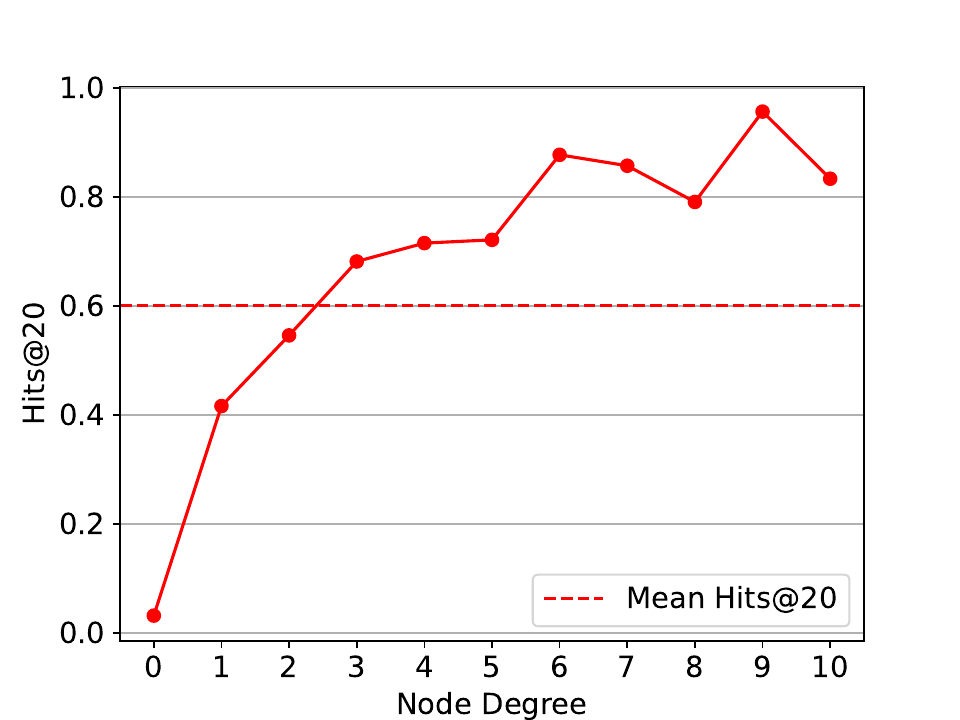}}  
    \subfigure[CiteSeer]{\includegraphics[width=0.325\linewidth]{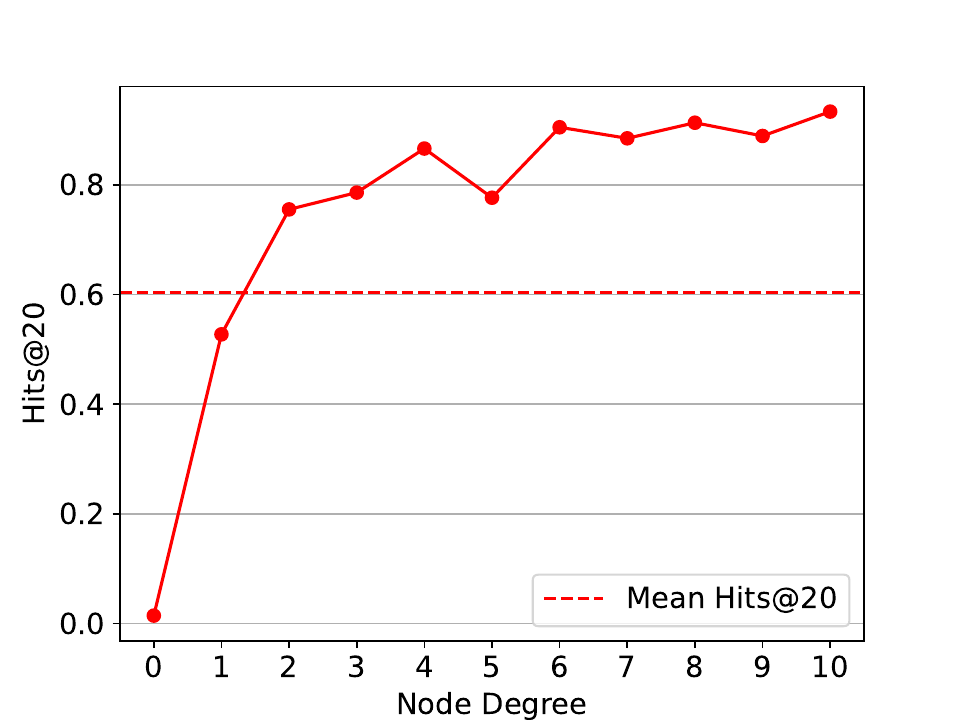}}
    \subfigure[CoraFull]{\includegraphics[width=0.325\linewidth]{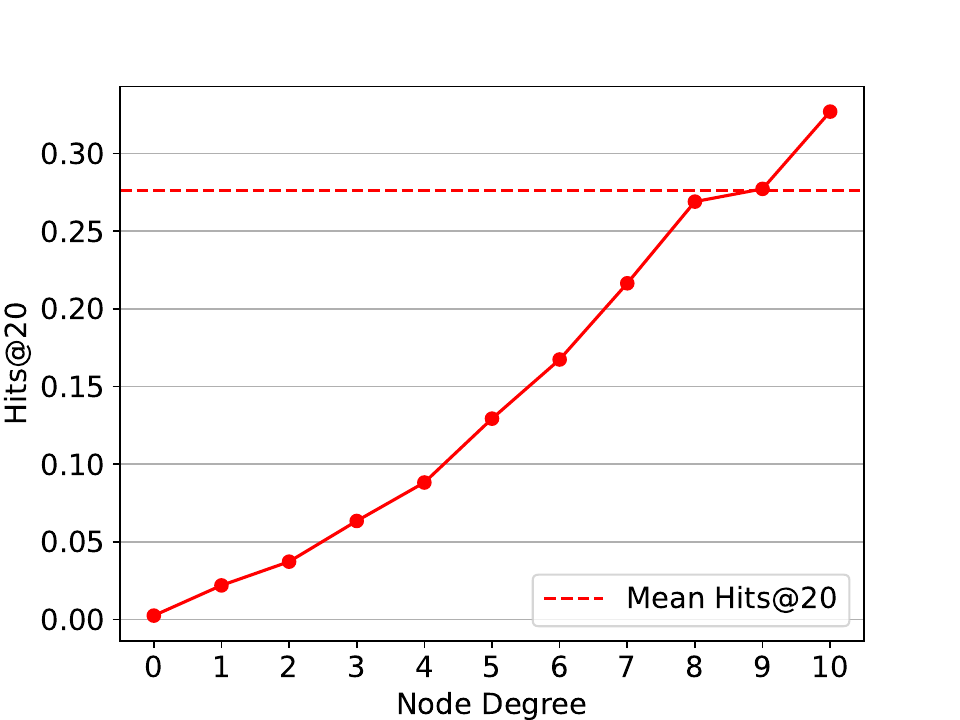}}
    \caption{GAE's LP performance distribution w.r.t. node degree.}
    \label{Fig: Performance Distribution}
\end{figure}

To investigate whether nodes with different degrees would have varying LP performance, we train GAE \cite{GAE} and report its Hits@20 score w.r.t. degree of nodes. From Fig. \ref{Fig: Performance Distribution}, we find that the LP performance also suffers heavily from the long-tailed node degree distribution. Specifically, nodes with low or zero degrees exhibit inferior LP performance compared to high-degree nodes. Such a performance disparity causes a performance bottleneck, given the characteristic long-tailed distribution of node degrees in real-world graphs.

\subsection{Varied Embedding Norms Across Node Degrees}\label{Sec: Embedding Norm Varying Node Degree}
\begin{figure}[h]
    \centering
    \subfigure[Cora]{\includegraphics[width=0.325\linewidth]{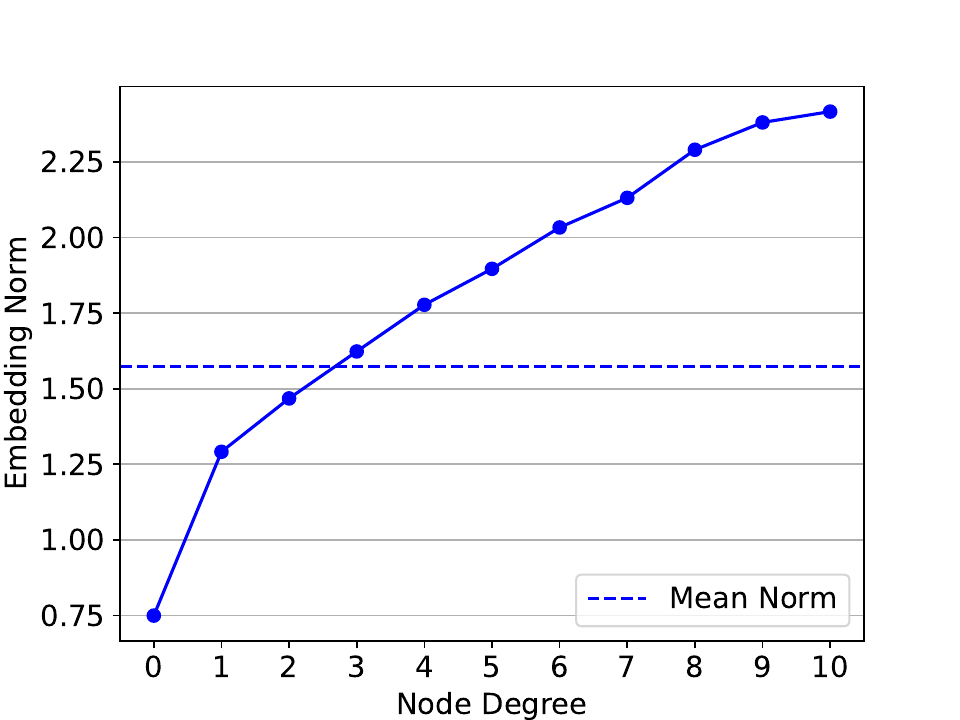}}  
    \subfigure[CiteSeer]{\includegraphics[width=0.325\linewidth]{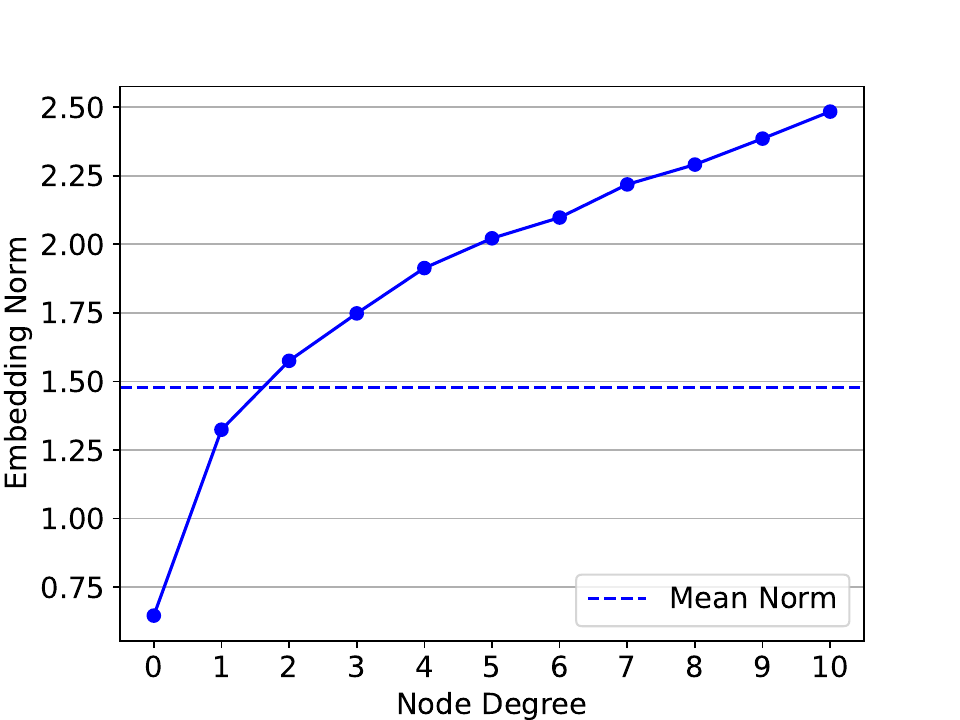}}
    \subfigure[CoraFull]{\includegraphics[width=0.325\linewidth]{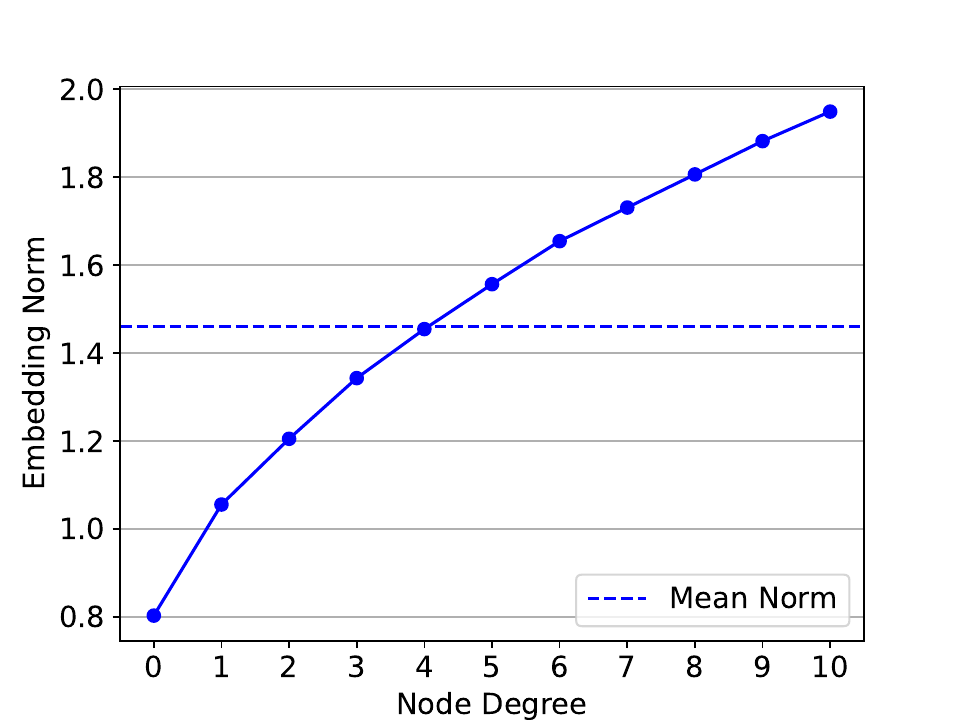}}
    \caption{Embedding norm distribution w.r.t. node degree.}
    \label{Fig: Norm Distribution}
\end{figure}

To identify the underlying cause of degree bias, previous studies \cite{SL-DSGCN, Tail-GNN, DegFairGNN} focusing on node classification hypothesize that neighborhoods of low-degree nodes contain insufficient or overly noisy information for effective representations. Moreover, they suggest that predictions for high-degree nodes are more susceptible to influence from training nodes due to their numerous links with other nodes. However, we find that the inherent unfairness within the training objective of LP plays a pivotal role in exacerbating degree-related performance disparities. Specifically, we can reformulate the positive term $\mathcal{L}^+$ of $\mathcal{L}_\text{GAEs}$ in a node-wise format as follows:
% In this work, we find that the inherent unfairness within the training objective of LP plays a pivotal role in exacerbating degree-related performance disparities. Specifically, we can reformulate the positive term $\mathcal{L}^+$ of $\mathcal{L}_\text{GAEs}$ in a node-wise format as follows:
\begin{equation}
    \mathcal{L}^+ = \frac{1}{|\mathcal{E}^+|} \sum_{v_i \in \mathcal{V}} \sum_{v_j \in \mathcal{N}_i}\log \sigma \left( ||\boldsymbol{z}_i|| \cdot ||\boldsymbol{z}_j|| \cdot \cos (\theta_{\boldsymbol{z}_i,\boldsymbol{z}_j}) \right),
\end{equation}
% \begin{equation}
% \begin{aligned}
%     \mathcal{L}^+ & = \frac{1}{|\mathcal{E}^+|}\sum_{(v_i, v_j)\in \mathcal{E}^+}\log \sigma(\boldsymbol{z}_i^\top \boldsymbol{z}_j) \\
%      & = \frac{1}{|\mathcal{E}^+|} \sum_{v_i \in \mathcal{V}} \sum_{v_j \in \mathcal{N}_i}\log \sigma(\boldsymbol{z}_{i}^\top \boldsymbol{z}_j) \\
%      & = \frac{1}{|\mathcal{E}^+|} \sum_{v_i \in \mathcal{V}} \sum_{v_j \in \mathcal{N}_i}\log \sigma \left( ||\boldsymbol{z}_i|| \cdot ||\boldsymbol{z}_j|| \cdot \cos (\theta_{\boldsymbol{z}_i,\boldsymbol{z}_j}) \right),
% \end{aligned}
% \end{equation}
where $||\boldsymbol{z}||$ denotes the $L_2$ norm of the embedding vector $\boldsymbol{z}$, and $\theta_{\boldsymbol{z}_i,\boldsymbol{z}_j}$ represents the angle between the two embeddings $\boldsymbol{z}_i,\boldsymbol{z}_j$. Optimizing this objective will pull linked nodes together by increasing the inner product of their embeddings. However, the sign of the inner product completely depends on the angle between two vectors. When the angle is smaller than $\frac{\pi}{2}$, then increasing the inner product can trivially become increasing the embedding norm. Consequently, high-degree nodes, which are related to more positive links, tend to have embeddings with larger norms. This observation is depicted in Fig. \ref{Fig: Norm Distribution}, where the mean norm of node embeddings learned by GAE is seen to be related to node degree.

\begin{figure}[h]
    \centering
    \subfigure[Cora]{\includegraphics[width=0.325\linewidth]{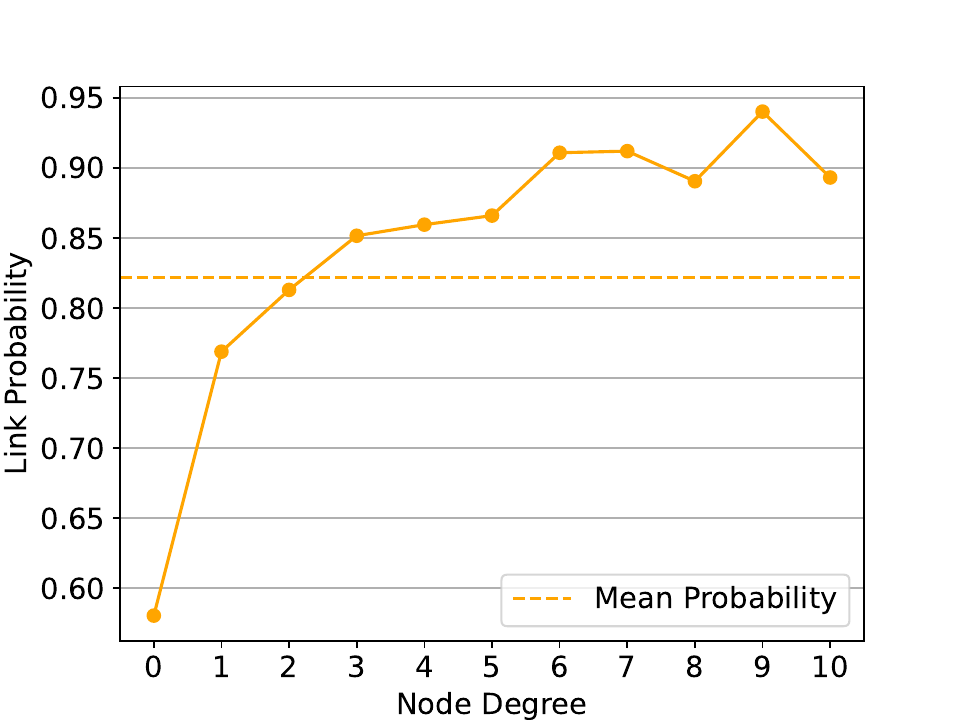}}  
    \subfigure[CiteSeer]{\includegraphics[width=0.325\linewidth]{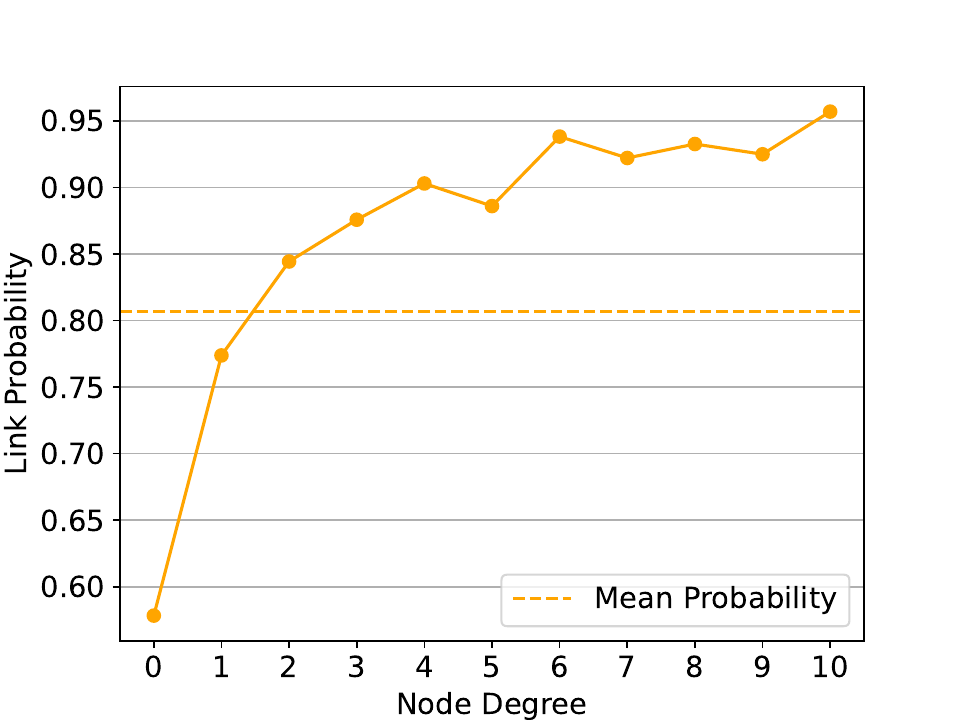}}
    \subfigure[CoraFull]{\includegraphics[width=0.325\linewidth]{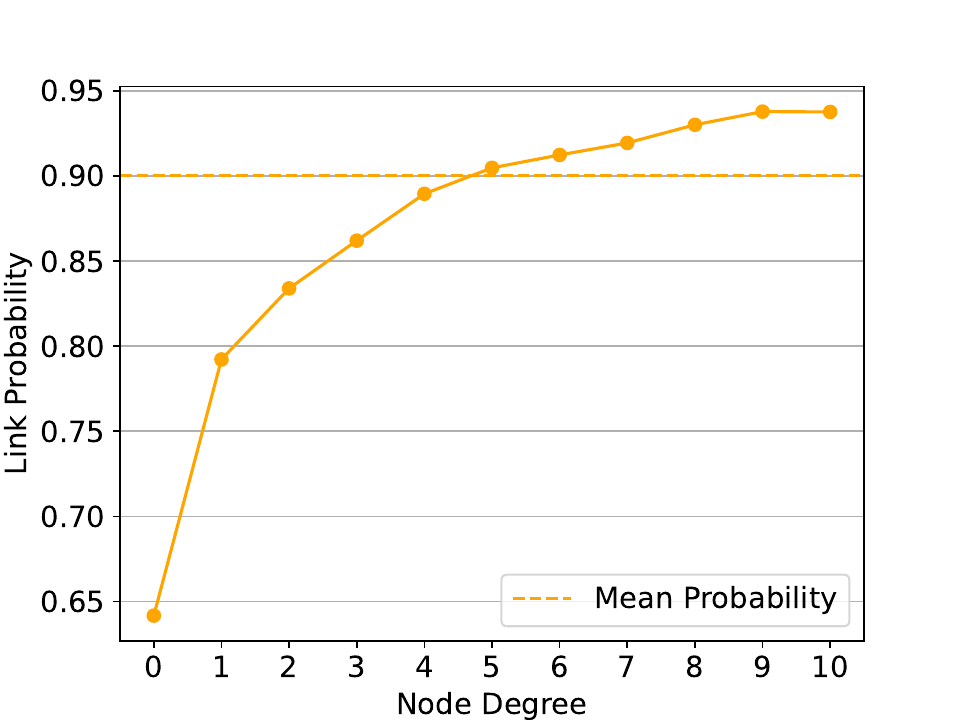}}
    \caption{Mean link probability of positive links in the test set w.r.t. node degree.}
    \label{Fig: Probability Distribution}
\end{figure}

During testing, given an anchor node $\boldsymbol{z}_i$ and a true linked node $\boldsymbol{z}_j$ from the test set, assuming the model accurately predicts the link's existence between them, i.e., $||\boldsymbol{z}_i|| \cdot ||\boldsymbol{z}_j|| \cdot \cos (\theta_{\boldsymbol{z}_i,\boldsymbol{z}_j}) > 0$, the larger the norm $||\boldsymbol{z}_j||$, the higher the predicted link probability $\sigma(\boldsymbol{z}_i^\top \boldsymbol{z}_j)$. In other words, embeddings with larger norms tend to yield higher probabilities for positive links, as depicted in Fig. \ref{Fig: Probability Distribution}. Similarly, larger norms also result in lower probabilities for accurately predicted negative links. As a result, higher link probabilities for positive links and lower link probabilities for negative links lead to superior performance scores.

\section{Proposed Method}
With the above analysis, we have identified norm bias as the primary factor underlying degree bias, which subsequently results in performance degradation. This indicates that nodes with higher degrees have larger embedding norms, resulting in imbalanced performance distribution over different degrees. Therefore, we propose increasing the embedding norms of low-degree nodes to mitigate the degree bias in GAEs. Given a predefined degree threshold $d_t$, we augment $(d_t-d_i)$ self-loops for each node $v_i$ with a degree $d_i$ lower than $d_t$. This embedding \textbf{N}orm \textbf{A}ugmentation (\textbf{NA}) strategy exploits the relationship where the square of an embedding's norm equals the inner product of itself, i.e., $||\boldsymbol{z_i}||^2 = \boldsymbol{z}_i^\top \boldsymbol{z}_i$. Hence, additional self-loops are introduced as positive samples into the training objective for low-degree nodes. Formally, the augmented loss function $\mathcal{L}^+$ is defined as:
\begin{equation}
    \mathcal{L}^+ = \frac{1}{m} \left( \sum_{(v_i, v_j)\in \mathcal{E}^+}\log \sigma(\boldsymbol{z}_i^\top \boldsymbol{z}_j) + \sum_{d_i < d_t}(d_t-d_i)\log \sigma(\boldsymbol{z}_i^\top \boldsymbol{z}_i) \right),
\end{equation}
where $m = |\mathcal{E}^+|+\sum_{d_i < d_t}(d_t-d_i)$. This straightforward strategy can be seamlessly integrated into various GAE models with minimal additional computational overhead, i.e., $\mathcal{O}(|\{v_i | d_i < d_t \}|)$. Despite its simplicity, it effectively mitigates the degree bias in GAEs (see Fig. \ref{Fig: Model Degree Fairness}). We note a similarity between our method of norm augmentation and the self-loop trick for training GNNs, such as the extra self-link in the normalized adjacency matrix of GCN \cite{GCN}, as they both introduce self-loop links. However, while self-loops in GNNs are used solely for aggregation purposes, the links introduced by our approach are used to increase embedding norms and can be seen as additional positive training samples for low-degree nodes.

\section{Experiments}
% In this section, we conduct experiments on real-world datasets to evaluate the effectiveness of NA. In particular, we aim to answer the following research questions:
% \begin{description}
%     \item[RQ1:] Can NA consistently boost the LP performance and mitigate the degree bias of different GAE frameworks?
    
%     \item[RQ2:] Does NA always outperform other methods designed to mitigate degree bias?

%     \item[RQ3:] Is the node embedding norm a critical factor in influencing the LP performance?

%     \item[RQ4:] How sensitive NA is to the degree threshold $d_t$, training ratio, and encoder architectures?
% \end{description}

\subsection{Experimental Setup}
% In this section, we conduct experiments on real-world datasets to evaluate the effectiveness of NA. In particular, we aim to answer the following research questions:
% \begin{description}
%     \item[RQ1:] Can NA consistently boost the LP performance and mitigate the degree bias of different GAE frameworks?
    
%     \item[RQ2:] Does NA always outperform other methods designed to mitigate degree bias?

%     \item[RQ4:] How sensitive NA is to the degree threshold $d_t$ and encoder architectures?
% \end{description}

\subsubsection{Datasets and Baselines}
% \noindent\textbf{Datasets.}
We conduct experiments on four citation graphs: Cora, CiteSeer, PubMed \cite{GAE}, and CoraFull \cite{G2G}. In these graphs, nodes represent papers, while edges denote citation links. The statistics of the datasets are shown in Table \ref{Tab: Dataset statistics}. Following \cite{LLP}, we randomly sample 5\%/15\% of the links with the same number of disconnected node pairs from the graph as the validation/test sets. And the validation/test links are excluded from the training graph.

We integrate NA with four GAE models for evaluation: GAE \cite{GAE}, ARGA \cite{ARGA}, LGAE \cite{LGAE}, and GNAE \cite{GNAE}. Additionally, we compare our method with three methods designed to alleviate degree bias: Tail-GNN \cite{Tail-GNN}, ResNorm \cite{ResNorm}, and HAW \cite{HAW}.

\begin{table}[h]
    \centering
    \caption{Dataset statistics.}
    \begin{tabular}{lcccc}
        \toprule
        Dataset      & \#Nodes    & \#Edges   & \#Features  & Mean Degree \\
        \midrule
        Cora         & 2,708       & 5,278     & 1,433        & 3.9         \\
        CiteSeer     & 3,327       & 4,552      & 3,703        & 2.7         \\
        PubMed       & 19,717      & 44,324     & 500         & 4.5         \\
        CoraFull     & 19,793      & 63,421     & 8,710        & 6.4         \\
        \bottomrule
    \end{tabular}
    \label{Tab: Dataset statistics}
\end{table}

\subsubsection{Evaluation and Implementation}
% \noindent\textbf{Evaluation.}
For all experiments, we use Hits@$K$ as the metric. Hits@$K$ ranks positive test links against negative test links and computes the ratio of positive test links ranked at $K$-th place or above. We report the averaged test performance along with its standard deviation over $10$ runs with different random initializations on fixed dataset splits. 

We employ PyG \cite{PyG} to implement all GAE models. All models are trained for a maximum of $500$ epochs using the Adam optimizer \cite{Adam} with a learning rate of $0.01$. We set the embedding dimension to 32 and the hidden dimension to 64 for all multi-layer GNNs. We choose the degree threshold $d_t$ from $\{ 2, 3, 4 \}$ based on the best validation performance.

\begin{table}[!ht]
\caption{Overall LP performance of the original and NA GAEs.}
    \label{Tab: GAEs'Link Prediction }
    \centering
    \begin{tabular}{lcccc}
    \toprule
        ~ & Cora & CiteSeer & PubMed & CoraFull \\ \midrule
        GAE & 59.34±0.81 & 60.64±1.27 & 37.04±1.14 & 27.74±1.09 \\
        \textbf{GAE-NA} & \textbf{70.56±1.37} & \textbf{78.83±0.74} & \textbf{40.91±0.66} & \textbf{29.29±0.82} \\ \midrule
        ARGA & 50.34±2.07 & 50.38±1.42 & 26.70±0.87 & 25.06±0.93 \\
        \textbf{ARGA-NA} & \textbf{57.85±1.51} & \textbf{62.99±0.70} & \textbf{28.30±1.07} & \textbf{26.01±0.38} \\ \midrule
        LGAE & 61.42±0.53 & 70.41±0.96 & 28.50±0.71 & 25.82±0.81 \\
        \textbf{LGAE-NA} & \textbf{67.35±0.94} & \textbf{84.34±0.79} & \textbf{31.32±1.20} & \textbf{26.13±0.82} \\ \midrule
        GNAE & 72.55±1.20 & 84.44±1.15 & 33.06±0.83 & 33.06±0.80 \\
        \textbf{GNAE-NA} & \textbf{73.23±0.89} & \textbf{86.04±0.96} & \textbf{34.78±0.46} & \textbf{33.73±1.04} \\
        \bottomrule
    \end{tabular}
\end{table}

\begin{figure}[!ht]
    \centering
    \subfigure[GAE-NA]{\includegraphics[width=0.24\linewidth]{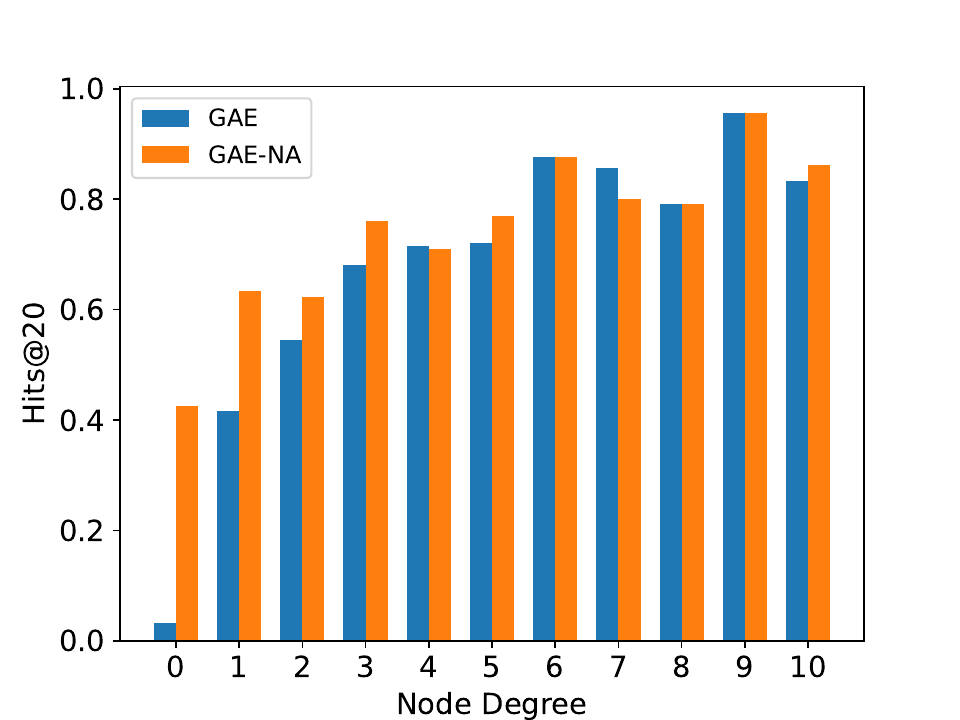}}  
    \subfigure[ARGA-NA]{\includegraphics[width=0.24\linewidth]{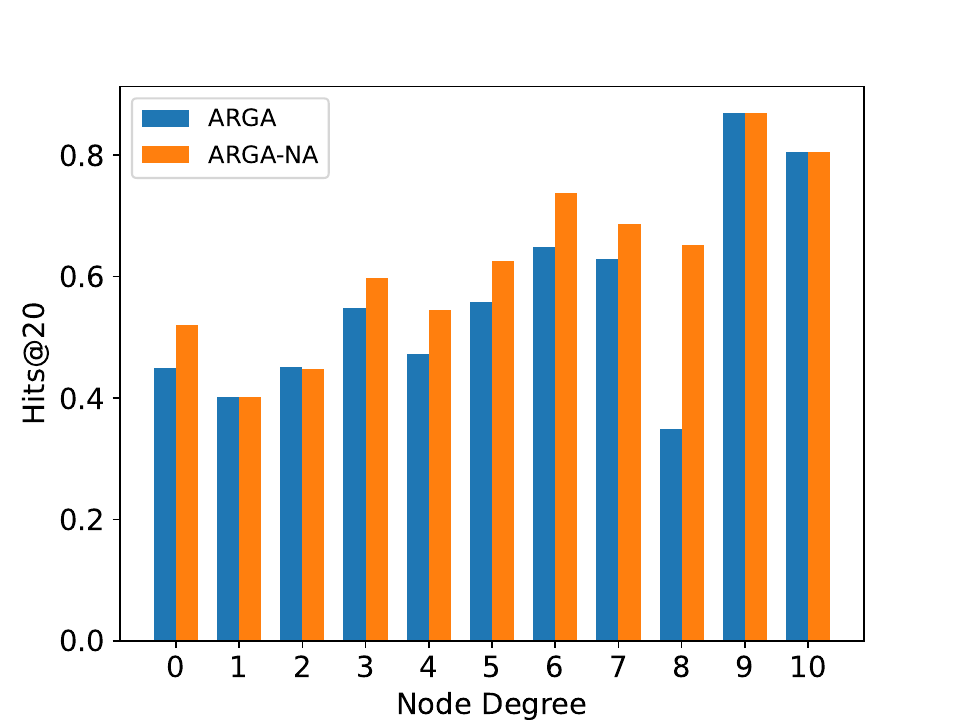}}
    \subfigure[LGAE-NA]{\includegraphics[width=0.24\linewidth]{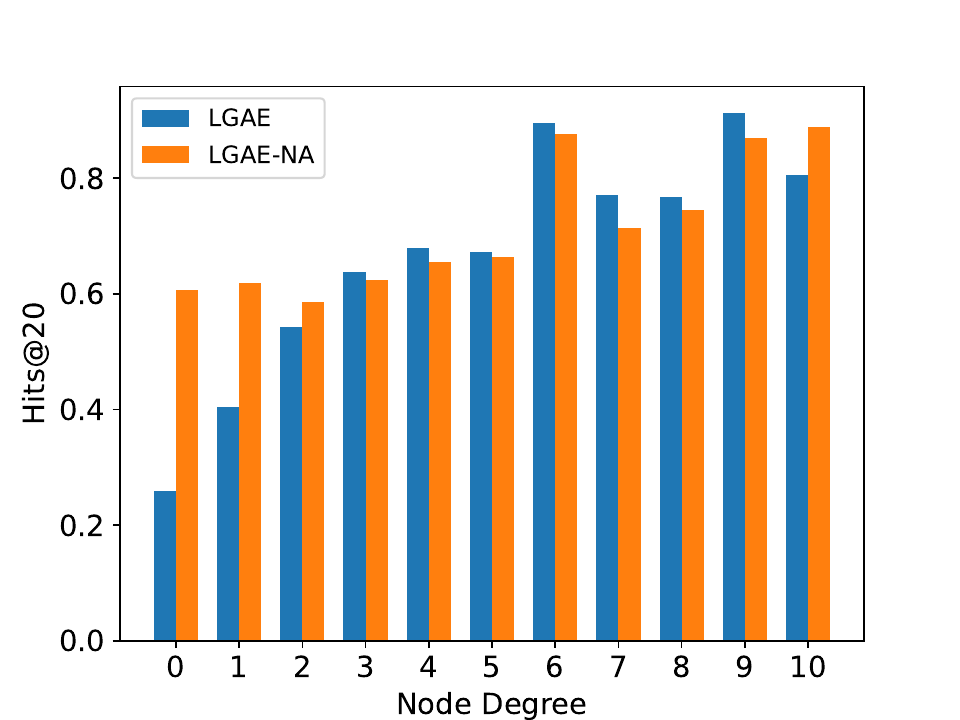}}
    \subfigure[GNAE-NA]{\includegraphics[width=0.24\linewidth]{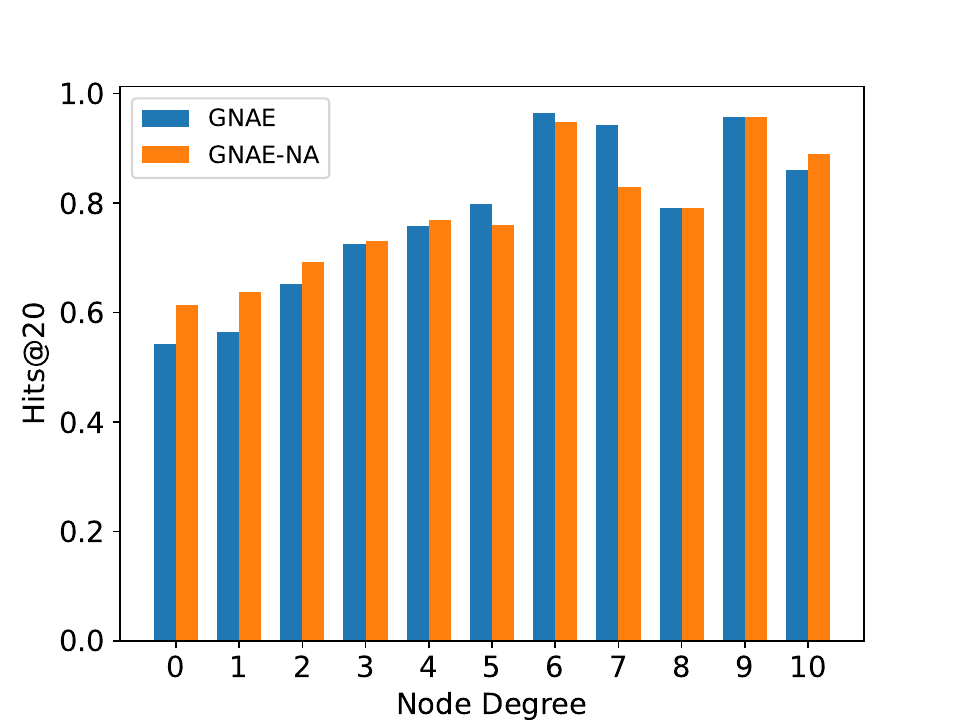}}
    \subfigure[GAE-NA]{\includegraphics[width=0.24\linewidth]{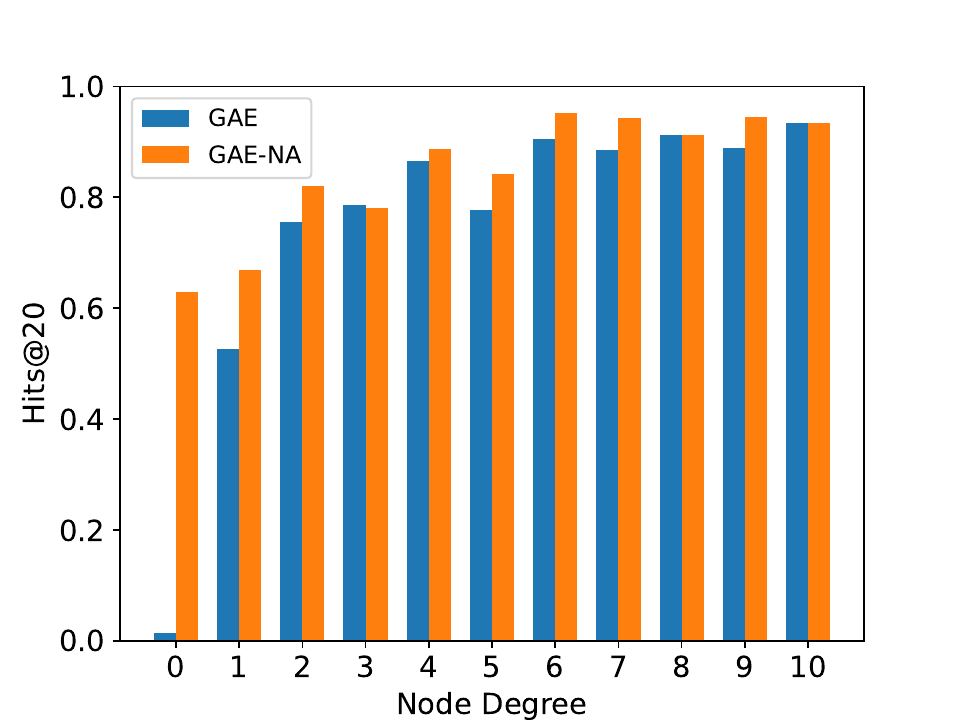}}  
    \subfigure[ARGA-NA]{\includegraphics[width=0.24\linewidth]{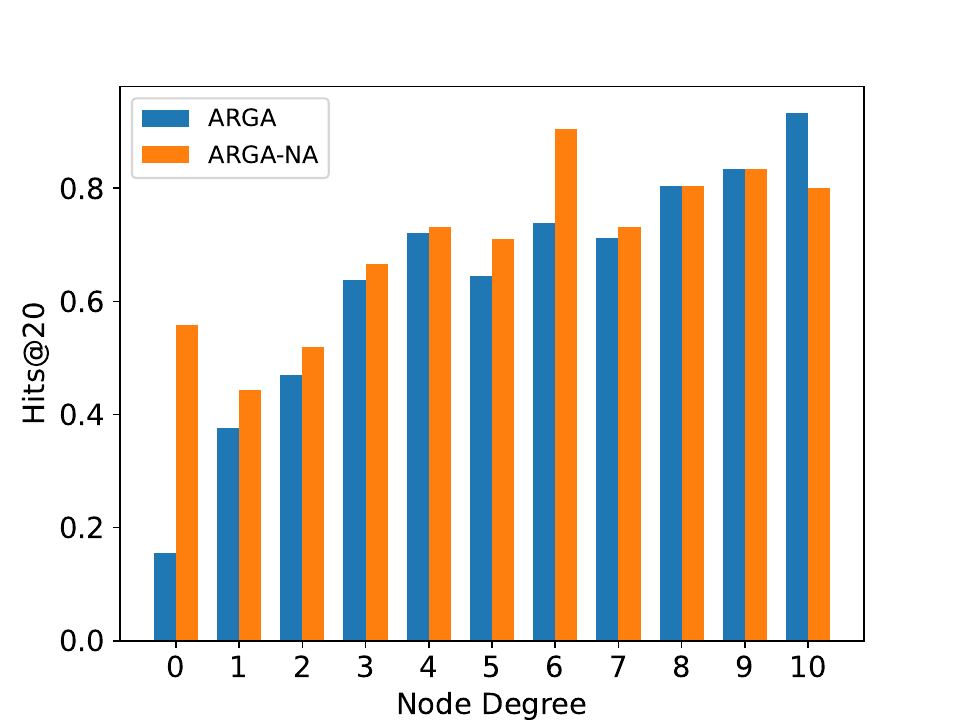}}
    \subfigure[LGAE-NA]{\includegraphics[width=0.24\linewidth]{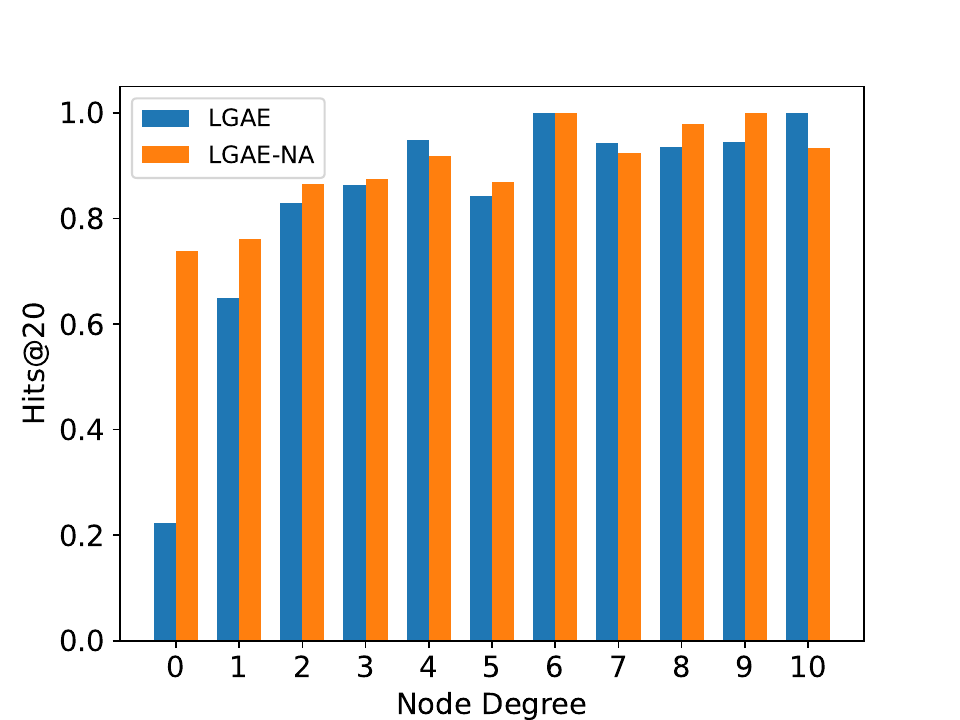}}
    \subfigure[GNAE-NA]{\includegraphics[width=0.24\linewidth]{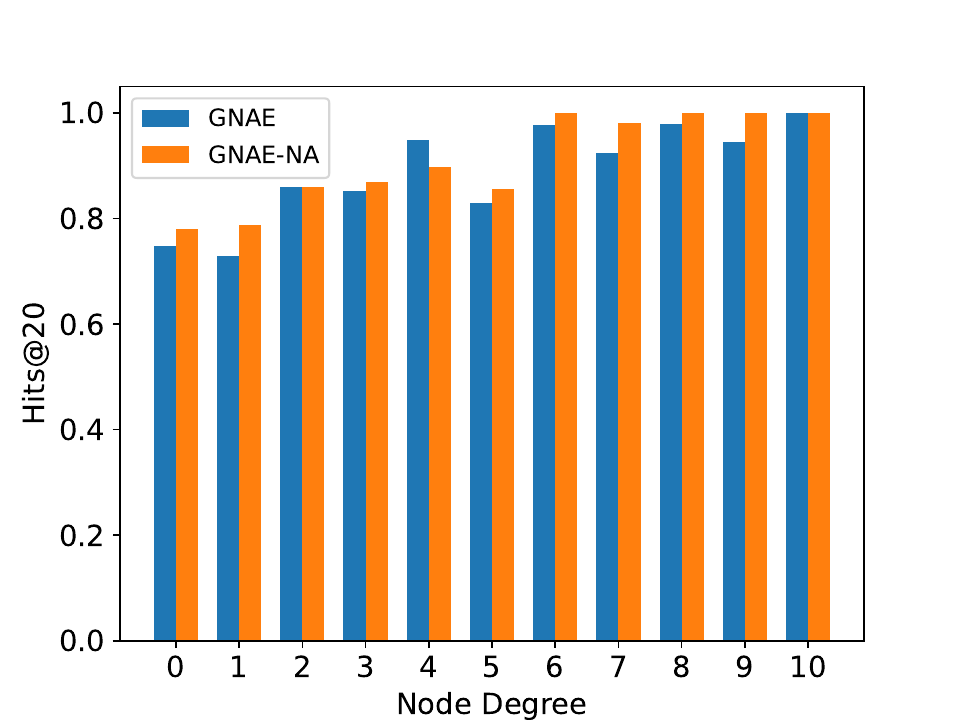}}
    \caption{LP performance of different norm-augmented GAEs on Cora (Top) and CiteSeer (Bottom) w.r.t. node degree.}
    \label{Fig: Cora CiteSeer GAE LP Performance}
\end{figure}

\subsection{Experimental Results}

\subsubsection{Comparison with Different GAE Backbones}
We integrate NA with four GAE backbones: GAE, ARGA, LGAE, and GNAE. Table \ref{Tab: GAEs'Link Prediction } shows the overall LP results. Generally, incorporating NA consistently improves the LP performance across different datasets and GAE frameworks. For instance, GAE-NA achieves performance gains by 11.22\%, 18.19\%, 3.87\%, and 1.55\% on the four datasets w.r.t. Hits@$20$, respectively. ARGA-NA outperforms ARGA by 7.51\%, and 12.61\% on Cora and CiteSeer, while LGAE-NA outperforms LGAE by 5.93\%, and 13.93\% on the two datasets, respectively. These results also suggest that NA can be successfully adopted in various GAE frameworks.

To validate the effectiveness of NA on low-degree nodes, we further visualize the LP performance of nodes with different degrees on Cora and Citeseer in Fig. \ref{Fig: Cora CiteSeer GAE LP Performance}. It can be observed that ARGA, LGAE, and GNAE exhibit a fairer treatment towards node degree compared to GAE, i.e., the performance disparities between high-degree and low-degree nodes in these three models are smaller than those in GAE. Our proposed NA further improves the LP performance of low-degree nodes (e.g., nodes with a degree smaller than $4$), while rarely having adverse effects on high-degree nodes. As a result, NA not only mitigates the degree bias in GAEs but also improves the overall performance significantly.

\begin{table}[!ht]
\caption{Overall LP performance compared with degree-fair methods.}
    \label{Tab: Degree-Fair Link Prediction}
    \centering
    \begin{tabular}{lcccc}
    \toprule
    ~ & Cora & CiteSeer & PubMed & CoraFull \\ \midrule
    GAE & 59.34±0.81 & 60.64±1.27 & 37.04±1.14 & 27.74±1.09 \\
    Tail-GNN & 61.54±1.34 & 67.39±1.86 & 35.37±0.80 & 26.55±0.83 \\
    ResNorm & 66.75±0.74 & 73.17±1.00 & 39.13±1.05 & 28.69±1.11 \\
    HAW & 62.80±1.09 & 70.03±1.03 & 32.35±1.00 & 24.43±1.17 \\
    \textbf{GAE-NA} & \textbf{70.56±1.37} & \textbf{78.83±0.74} & \textbf{40.91±0.66} & \textbf{29.29±0.82} \\ \midrule
    LGAE & 61.42±0.53 & 70.41±0.96 & 28.50±0.71 & 25.82±0.81 \\
    Tail-GNN & 62.53±0.80 & 73.52±0.82 & 25.29±0.82 & 25.33±0.51 \\
    ResNorm & 66.73±0.49 & 82.26±0.78 & 29.28±0.46 & \textbf{26.41±1.26} \\
    HAW & 64.35±0.64 & 75.87±0.89 & 28.12±0.87 & 22.31±0.60 \\
    \textbf{LGAE-NA} & \textbf{67.35±0.94} & \textbf{84.34±0.79} & \textbf{31.32±1.20} & 26.13±0.82 \\
    \bottomrule
    \end{tabular}
\end{table}

\begin{figure}[h]
    \centering
    \subfigure[GAE]{\includegraphics[width=0.24\linewidth]{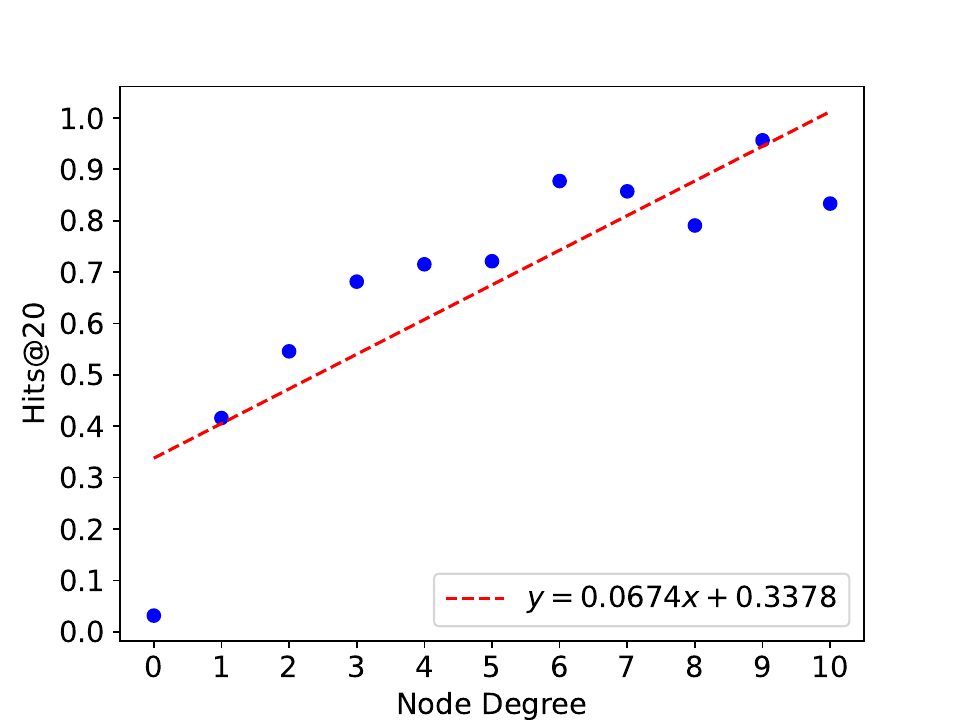}}  
    \subfigure[ResNorm]{\includegraphics[width=0.24\linewidth]{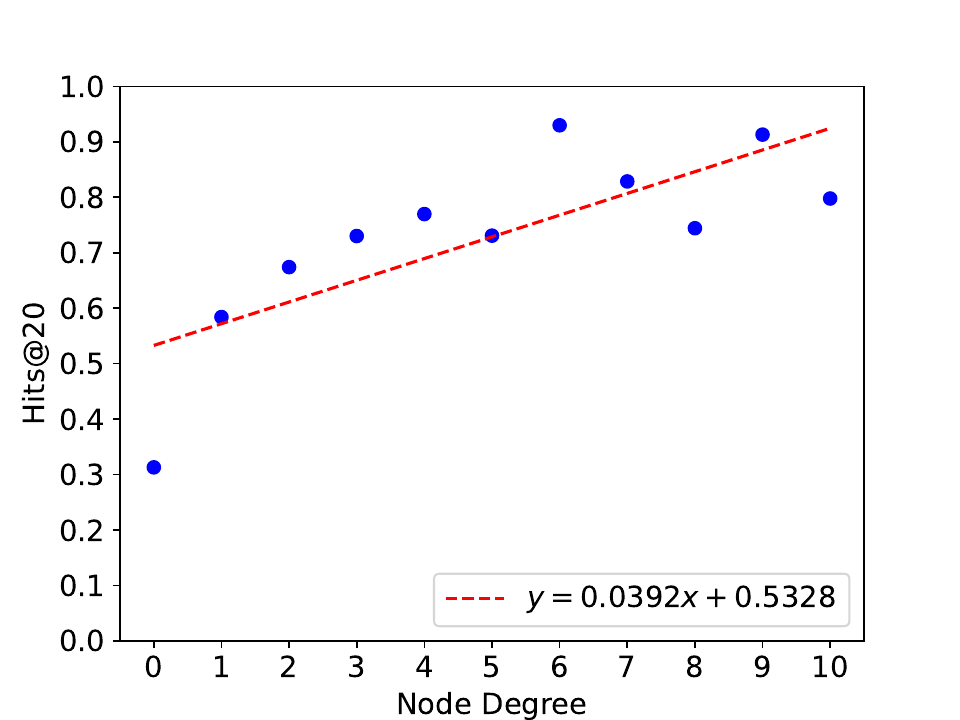}}
    \subfigure[HAW]{\includegraphics[width=0.24\linewidth]{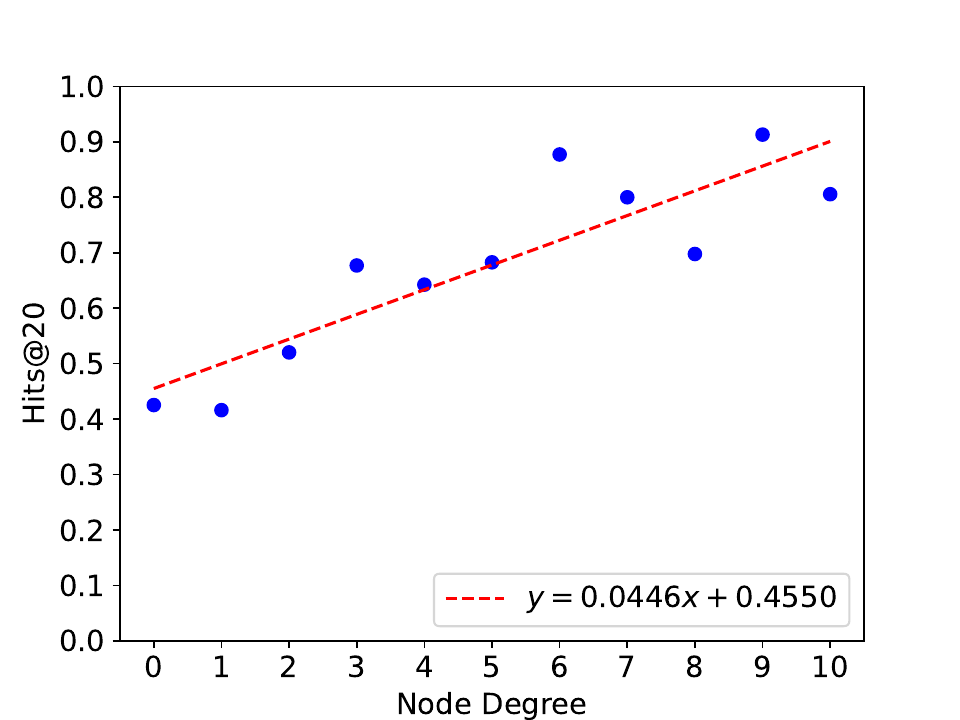}}
    \subfigure[GAE-NA]{\includegraphics[width=0.24\linewidth]{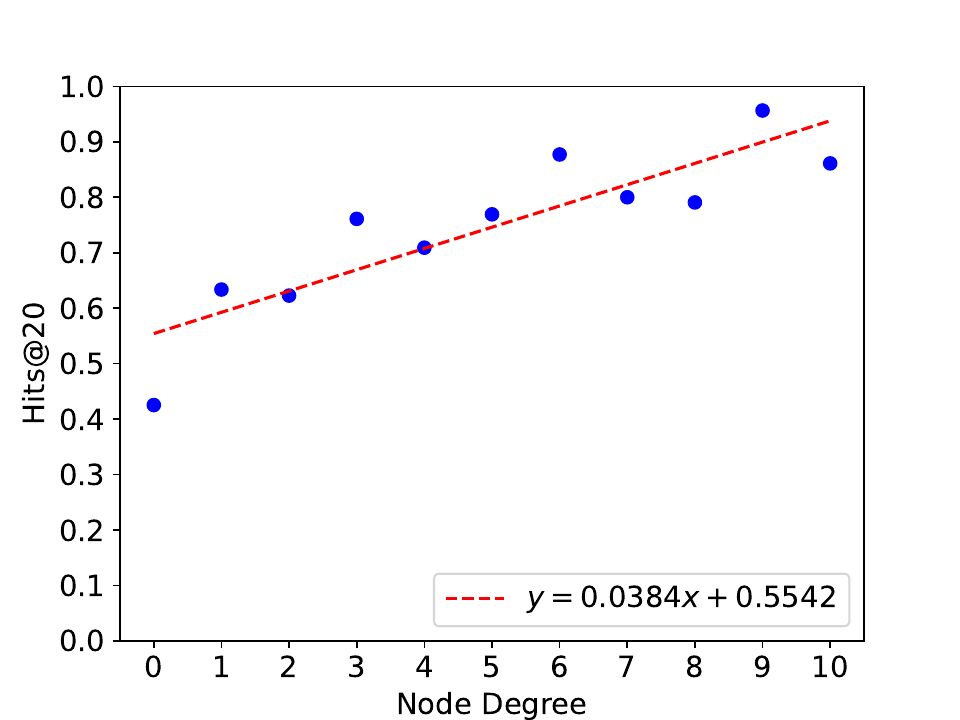}}
    \subfigure[GAE]{\includegraphics[width=0.24\linewidth]{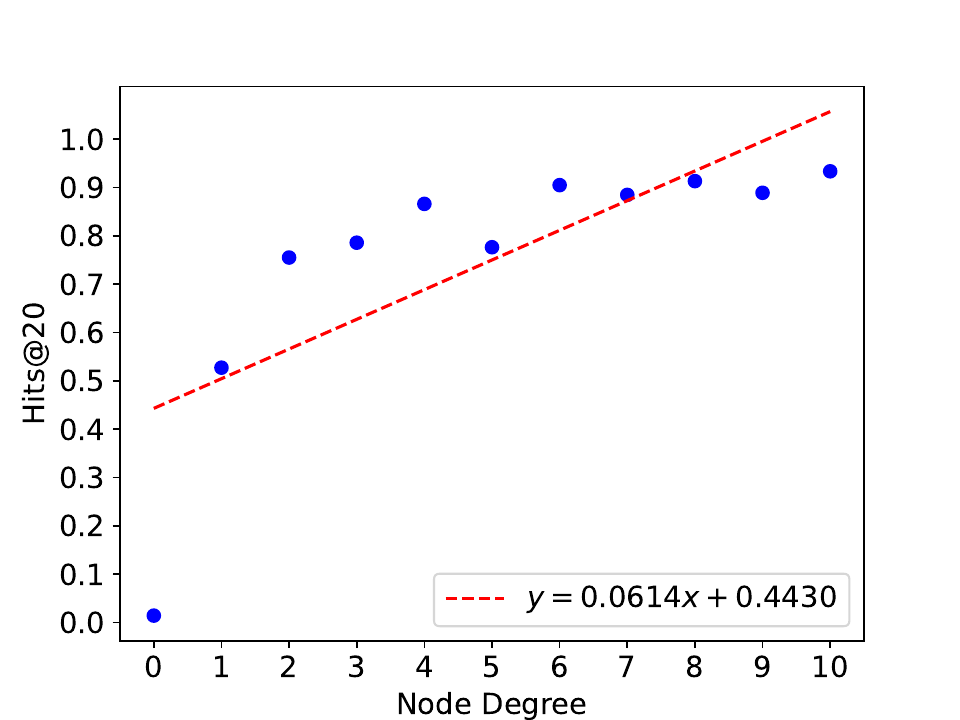}}  
    \subfigure[ResNorm]{\includegraphics[width=0.24\linewidth]{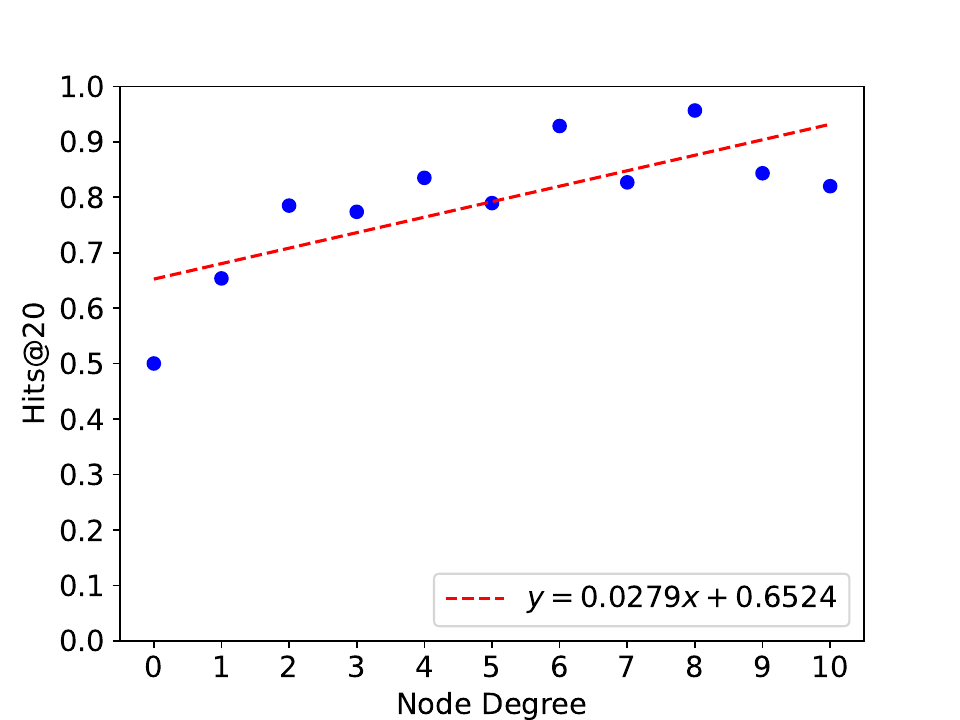}}
    \subfigure[HAW]{\includegraphics[width=0.24\linewidth]{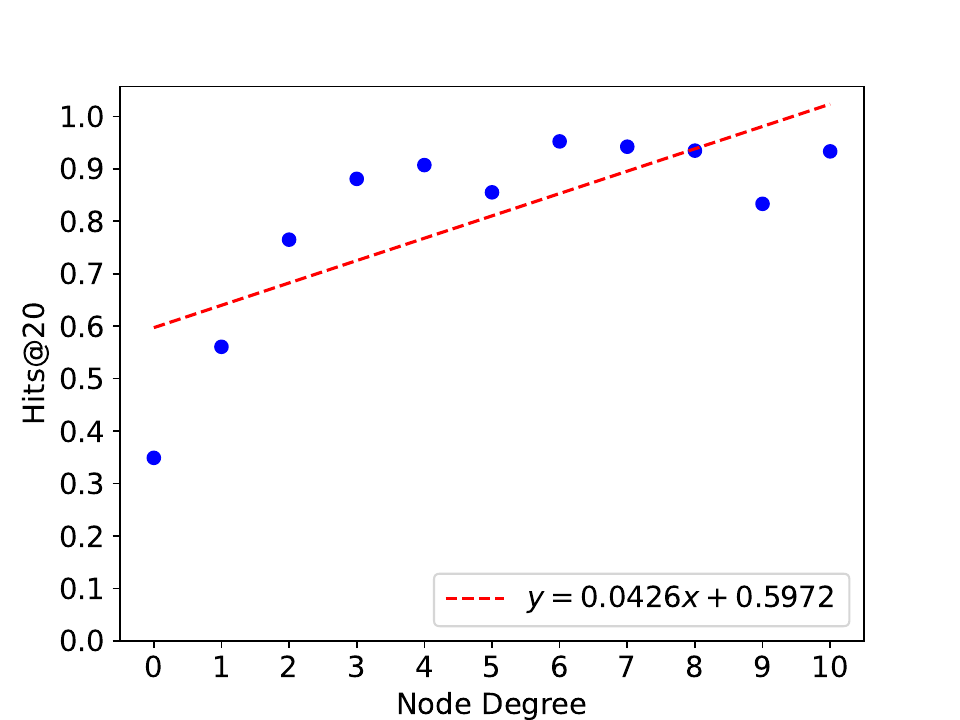}}
    \subfigure[GAE-NA]{\includegraphics[width=0.24\linewidth]{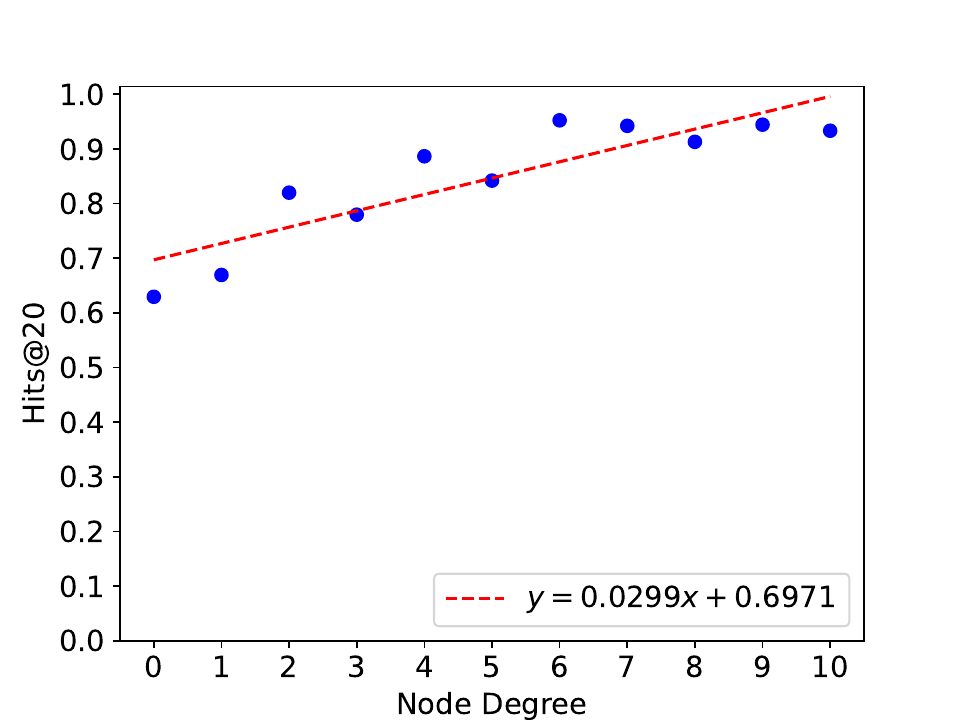}}
    \caption{Visualization for the fairness of models to degree bias. Blue scatters refer to the average Hits@$20$ of nodes with a specific degree. Red dotted lines are regression lines of blue scatters in each figure. For clarity, we annotate analytical expressions of regression lines. The higher the average Hits@$20$ of low-degree nodes and the smaller the slope of the regression line indicate that the model is fairer to degree bias.}
    \label{Fig: Model Degree Fairness}
\end{figure}

\subsubsection{Comparison with Degree-Fair Methods}
We additionally compare our NA with other degree-fair methods, including Tail-GNN, ResNorm, and HAW, utilizing the GAE and LGAE frameworks. The overall results on four datasets are shown in Table \ref{Tab: Degree-Fair Link Prediction}. It is observed that NA outperforms all degree-fair baselines on three datasets, except for CoraFull, where NA also achieves the runner-up performance. Furthermore, it is noteworthy that Tail-GNN and HAW do not consistently improve over GAE/LGAE; for instance, they degrade performance on the PubMed and CoraFull datasets.

To illustrate their fairness to degree bias, we then plot the LP performance of GAE, ResNorm, HAW, and GAE-NA w.r.t. node degree on the Cora and CiteSeer datasets, as shown in Fig. \ref{Fig: Model Degree Fairness}. Additionally, we conducted a linear regression analysis to visualize how these models balance performance between low-degree and high-degree nodes. A flat slope in the regression line indicates fairness to degree bias. From the figure, we can find that the Hits@$20$ for low-degree nodes of ResNorm, HAW and GAE-NA is higher than that of GAE, and the slope of the regression line is also smaller. These observations suggest that these methods are fairer than GAE under degree bias. Moreover, NA and ResNorm exhibit comparable degree fairness.

\subsubsection{Sensitivity Analysis}

\begin{figure}
    \centering
    \subfigure[Cora]{\includegraphics[width=0.49\linewidth]{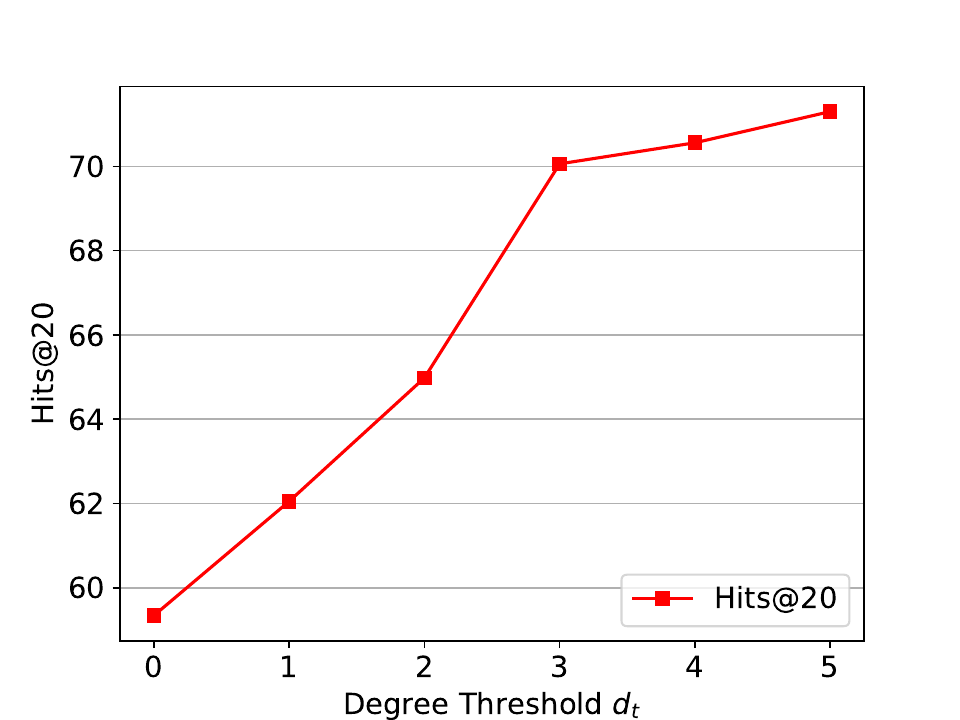}}  
    \subfigure[CiteSeer]{\includegraphics[width=0.49 \linewidth]{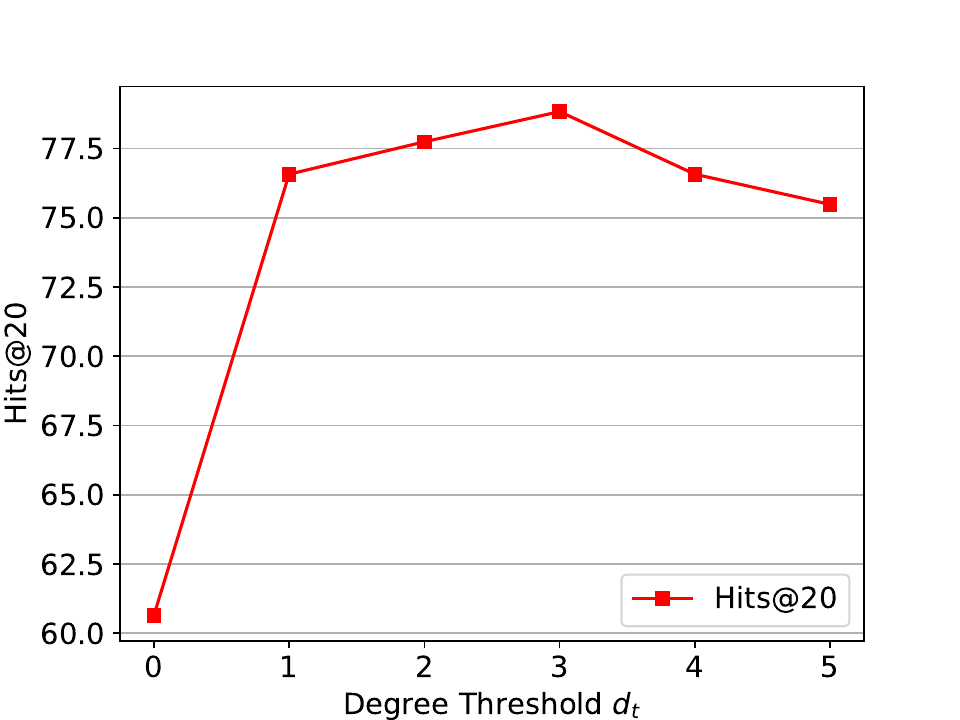}}
    \caption{LP performance of GAE-NA with different degree thresholds.}
    \label{Fig: Degree Threshold}
\end{figure}

\noindent\textbf{Selection of the Degree Threshold $d_t$.}
We add $(d_t-d_i)$ self-loops for each node with a degree $d_i$ less than $d_t$. To evaluate the impact of different thresholds, we report the results on the Cora and CiteSeer datasets in Fig. \ref{Fig: Degree Threshold}. It can be observed that integrating NA ($d_t >= 1$) consistently improves the LP performance of GAE ($d_t=0$), especially in terms of Hits@$20$. Determining the degree threshold can be guided by the data-driven analysis, as illustrated in Fig. \ref{Fig: Norm Distribution}. For instance, on CiteSeer, we set $d_t$ to $2$ since nodes with degrees not exceeding $2$ manifest below-average embedding norms.

\begin{figure}
    \centering
    \subfigure[Cora]{\includegraphics[width=0.49\linewidth]{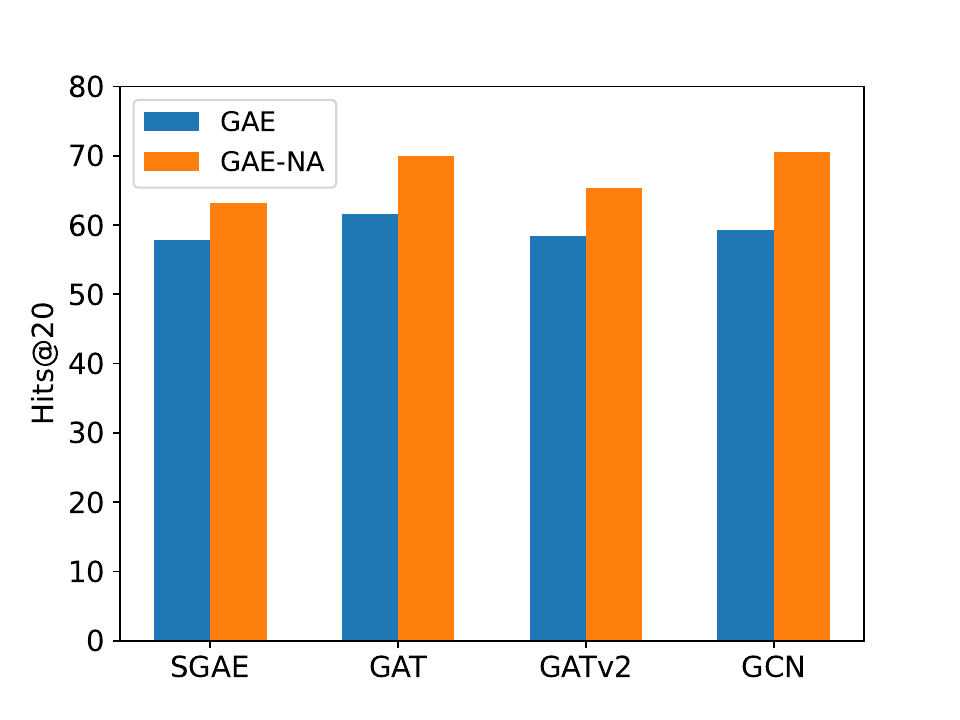}}  
    \subfigure[CiteSeer]{\includegraphics[width=0.49 \linewidth]{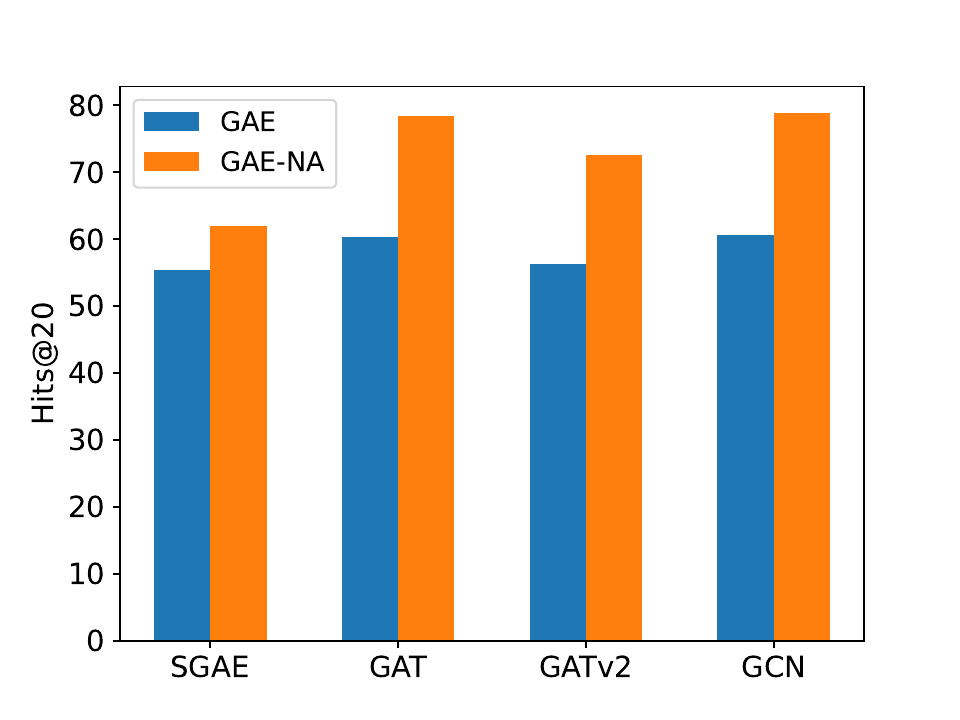}}
    \caption{LP performance of GAE-NA with different graph encoders.}
    \label{Fig: Different Encoder}
\end{figure}

\noindent\textbf{Robustness to Different Encoders.}
In previous experiments, GAE is evaluated using GCN as the encoder. Here, we extend the evaluation of GAE and GAE-NA, employing other popular graph encoders: SGAE \cite{SAGE}, GAT \cite{GAT}, and GATv2 \cite{GATv2}. Fig. \ref{Fig: Different Encoder} illustrates the results, indicating that our proposed NA significantly enhances the overall performance of GAE across different encoders on Cora and CiteSeer datasets. This observation suggests that NA is a robust and effective strategy to enhance GAE.

\section{Conclusion}
In this study, we shed light on the degree-related bias in GAEs, which suggests that low-degree nodes tend to exhibit inferior LP performance compared to high-degree nodes. We observe the norm of node embeddings learned by GAEs is related to node degree and provide explanations for why high-degree nodes commonly have larger embedding norms. This norm bias bridges degree bias and performance bias. Therefore, we propose to improve GAEs’ LP performance on low-degree nodes by increasing their embedding norms, which is simply implemented by introducing additional self-loops into the training objective for low-degree nodes. Experimental results confirm the effectiveness of our norm augmentation strategy across various scenarios.

\section*{Acknowledgment}
This work is partially supported by the National Key Research and Development Program of China (2021YFB1715600), and the National Natural Science Foundation of China (62306137).

% \clearpage
\bibliographystyle{IEEEtran}
\bibliography{main}

\end{document}